\documentclass[11pt, letterpaper]{article}

\usepackage[utf8]{inputenc}
\usepackage{geometry}
\usepackage{graphicx} 
\usepackage{booktabs} 
\usepackage{authblk}  
\usepackage{hyperref} 
\usepackage{natbib}   
\usepackage{times}    
\usepackage{url}      
\usepackage{float}    
\usepackage{placeins} 

\title{\textbf{Is Sanskrit the Most Token-Efficient Language?\\A Quantitative Study using GPT, Gemini, and SentencePiece}}

\author{Anshul Kumar}
\affil{Independent Researcher, BITS Pilani Goa (Alumnus) \\ \texttt{f20190651g@alumni.bits-pilani.ac.in}}

\date{} 

\begin{document}

\maketitle

\begin{abstract}
Tokens are the basic units of Large Language Models (LLMs). LLMs rely on tokenizers to segment text into these tokens, and tokenization is the primary determinant of computational and inference cost. Sanskrit, one of the oldest languages, is hypothesized to express more meaning per token due to its morphology and grammar rules; however, no prior work has quantified this. We use a dataset of 701 parallel verses of the Bhagavad Gita, which comprises three languages—Sanskrit, English, and Hindi—along with transliteration of Sanskrit into English. We test tokenizers including SentencePiece (SPM), older GPT models, and the latest generation tokenizers from Gemini and GPT. We use metrics of token count, characters per token (token efficiency), and tokens per character (token cost). Results show a $\sim$2x difference in token counts between Sanskrit and English/Hindi under the unbiased SPM baseline. English/Hindi translations of Sanskrit commentary resulted in an approximately 20x increase in token count. GPT o200k\_base (latest, used by GPT-4o) and Gemini (latest) reduce bias by a significant degree compared to GPT cl100k\_base (used until GPT-4), but still fail to fully capture Sanskrit's compactness. This matters because there might be a penalty bias for non-English users, which inflates the token count. This research provides a foundation for improving future tokenizer design and shows the potential of Sanskrit for highly compact encoding, saving on cost while speeding up training and inference. The code and dataset are available at \url{https://github.com/anshulkr713/sanskrit-token-efficiency}.
\end{abstract}

\section{Introduction}
Large Language Models (LLMs) do not process text; they process tokens. The number of tokens produced by a tokenizer directly influences training cost, inference latency, memory use, and billing. However, since most tokenizers are trained primarily on English corpora, their behavior varies for different languages. This may result in disparity in efficiency and cost depending on the user’s language, functioning as a “token tax” and raising concerns over input language fairness.

Sanskrit is a highly information-dense language because of its rich morphological structure, case endings (\textit{Vibhakti}), and rigid grammar rules. These features allow Sanskrit to encode semantic relationships within words rather than relying on specific function words or strict word order. For example, where English may require multiple words to represent an action and its participants, Sanskrit can encode complex semantic relationships within a single word. Theoretically, this should result in a lower token count and usage. However, despite this high information density, there is little empirical work quantifying the compactness of Sanskrit using modern tokenizers.

To fill this gap, we form a parallel multilingual dataset of 701 verses from the Bhagavad Gita containing verses in Sanskrit, translations in English and Hindi, along with Sanskrit transliteration. This corpus provides a controlled setting for comparing how similar semantic content is segmented across languages and for measuring the compactness of the Sanskrit text relative to its translations and commentaries.

For this research, we evaluate four different tokenization systems: OpenAI’s legacy tokenizer (\texttt{cl100k\_base}), the latest GPT tokenizer (\texttt{o200k\_base}), Google’s Gemini tokenizer, and SentencePiece BPE models trained separately for Sanskrit, English, and Hindi with equal vocabulary sizes of around 8k to avoid vocabulary bias. To observe multi-language efficiency, we measure token count, characters per token (information density), and tokens per character (representation cost).

This work captures four major contributions:
\begin{enumerate}
    \item It presents the quantitative measurement of Sanskrit token efficiency relative to English and Hindi.
    \item It benchmarks the token bias across major tokenizers like \texttt{cl100k\_base}, \texttt{o200k\_base}, and Gemini for multi-language inputs.
    \item It shows that while SentencePiece models capture Sanskrit’s compactness, production-level state-of-the-art tokenizers still struggle to do so.
    \item It highlights the economic and fairness implications of these findings for the deployment of multilingual AI.
\end{enumerate}

\section{Related Work}
Tokenization following the subword method is the standard input representation for LLMs, replacing character and word-level encodings. Byte-Pair Encoding (BPE) and WordPiece \citep{sennrich2016neural, nakajima2012japanese} segment input text into subwords to balance the size of the vocabulary. SentencePiece \citep{kudo2018sentencepiece} introduced a new language-agnostic training and tokenization method on raw character sequences, not relying on whitespaces. These are some of the methods adopted in today’s LLMs; however, their behavior can vary substantially across multiple languages depending on the training corpus.

Recent work shows disparity in tokenization efficiency across languages. \citet{tan2024multilingual} showed that LLMs require more tokens to represent morphologically rich languages compared to English, influencing both inference cost and accuracy. \citet{ahia2023do} showed that high-resource languages dominate the vocabularies of tokenizers, which leads to disproportionate token sequence lengths for multilingual users. Also, \citet{dodge2021documenting} and \citet{wei2023impact} further showed that the choice of tokenizer can increase performance gaps across languages when the model architecture is kept the same.

In the case of Indic languages, many studies focus on morphology and computational representation. \citet{hellwig2018dcs} produced a huge annotated corpus of Sanskrit, and \citet{krishna2020neural} developed neural network-based models for Sandhi splitting and morphological tagging. \citet{kulkarni2022sanskrit} improved the segmentation of Sanskrit using transformer-based architectures. However, these works focus on modeling rather than comparing tokenization efficiency.

Comparative evaluations of Sanskrit vs. English and Hindi are limited. Most studies exclude Sanskrit entirely due to a lack of clean and valid datasets, as Sanskrit is a low-resource language. No prior study has examined how production-level tokenizers segment Sanskrit text relative to equivalent semantic content in Hindi and English. This absence leaves a gap in research regarding the hypothesized semantic compactness of Sanskrit. Quantifying token efficiency is important for assessing multilingual accuracy, reducing language bias, managing economic costs in API-based LLM usage, and informing future tokenizer designs. This study aims to fill this gap through a comparison of Sanskrit, Sanskrit transliteration, English, and Hindi using major tokenizers.

\section{Dataset}
We use an open-source dataset of the Bhagavad Gita containing 701 verses. The corpus includes the original Sanskrit verses (in Devanagari script), Latin transliteration of the same verse, and corresponding translations in English and Hindi. Also, the dataset distinguishes between the translation of the verse and the commentary (meaning). This distinction gives us the freedom to evaluate tokenization efficiency on both verse text and descriptive commentary.

Before analysis, we performed minimal preprocessing to ensure consistency across the records of the dataset. This involved removing HTML tags, cleaning punctuation, and normalizing all text to Unicode UTF-8. We did not alter the semantic content of the verses. As the dataset is open-source, it allows for reproducibility and experiments without copyright issues.

\section{Tokenizers and Evaluation Metrics}
We evaluate four tokenization systems to measure the efficiency of tokens across English, Hindi, and Sanskrit (Devanagari and Latin script). We focus on three production-level tokenizers actively used by major labs:
\begin{itemize}
    \item \textbf{\texttt{cl100k\_base}:} The legacy tokenizer for OpenAI GPT-3 and GPT-4 models.
    \item \textbf{\texttt{o200k\_base}:} The tokenizer used in the latest models (GPT-4o), designed to improve multilingual compression and reduce punctuation fragmentation compared to its previous version.
    \item \textbf{Gemini:} The tokenizer utilized by the Gemini 1.5 model family.
\end{itemize}

Along with these tokenizers, we include a controlled baseline using SentencePiece BPE. We train three different SentencePiece models separately for Sanskrit, English, and Hindi with a fixed vocabulary size of 8k tokens to avoid language bias, since Sanskrit has a larger vocabulary compared to the other two languages. Moreover, SentencePiece BPE runs directly on Unicode characters and is whitespace-agnostic, making it suitable for morphologically rich languages like Sanskrit.

We compute three evaluation metrics to assess cross-language efficiency:
\begin{enumerate}
    \item \textbf{Token Count:} The total tokens required to represent a text. LLM training and inference costs scale with sequence length, making this a direct factor for economic and computational cost.
    \item \textbf{Characters per Token (CpT):} A metric for \textbf{information density}. It shows the average amount of text captured by a single token; a large value indicates higher compactness.
    \item \textbf{Tokens per Character (TpC):} A metric for \textbf{fragmentation}. It quantifies the cost of representing text; lower values indicate that the tokenizer is efficiently utilizing its vocabulary to represent the language.
\end{enumerate}

\section{Experiments Section}
We conducted two experiments in a controlled environment to evaluate the behavior of tokenizers on the dataset. Both experiments were conducted on the same set of 701 Bhagavad Gita verses but used different input configurations to test distinct hypotheses.

\subsection{Experiment 1: Parallel Translations}
In this experiment, we evaluated the tokenizers on each language input to measure cross-language efficiency for equivalent semantic content. For each verse, four inputs were provided:
\begin{enumerate}
    \item Sanskrit (Devanagari script)
    \item Sanskrit Transliteration (Latin script)
    \item English (Translation)
    \item Hindi (Translation)
\end{enumerate}
For every verse, we recorded the token count, characters per token, and tokens per character without applying prompt formatting.

\subsection{Experiment 2: Semantic Expansion}
This experiment measures the degree of \textbf{semantic expansion}, meaning the amount of additional text required to unpack the meaning of a Sanskrit verse into other languages. For each verse, we tokenized the following segments independently:
\begin{enumerate}
    \item \textbf{Sanskrit Verse:} Original compact text in Devanagari.
    \item \textbf{English Translation:} Direct translation of the verse.
    \item \textbf{English Meaning:} Extended commentary/purport.
    \item \textbf{Hindi Translation:} Direct translation of the verse.
    \item \textbf{Hindi Meaning:} Extended commentary/purport.
\end{enumerate}
We used the same tokenizers from Experiment 1. This method allows us to quantify precisely how much the token count expands when moving from the dense structure of Sanskrit to the explanatory prose required in English and Hindi.

\paragraph{Implementation Details}
In both experiments, tokenizers were applied without any model inference. This ensures the results are based on the tokenizer's vocabulary behavior rather than the model's predictive properties. All calculations were performed programmatically, and results were exported to CSV files for analysis.

\section{Results}

\subsection{Experiment 1: Language Efficiency \& Density}
The table below presents the efficiency of each tokenizer. We report Mean Token Count and Characters per Token (CpT) for information density.

\begin{itemize}
    \item \textbf{Key Finding:} Under the SentencePiece (SPM) baseline, Sanskrit demonstrates superior information density (5.07 chars/token) and requires significantly fewer tokens (21.67) than English (41.96).
    \item \textbf{Legacy Bias:} The \texttt{cl100k\_base} model fails to capture this density, forcing Sanskrit’s CpT down to 1.13.
\end{itemize}

\begin{table}[ht]
\centering
\caption{Mean Token Count \& Density (Experiment 1)}
\label{tab:exp1}
\resizebox{\textwidth}{!}{%
\begin{tabular}{@{}lccccc@{}}
\toprule
\textbf{Tokenizer} & \textbf{English} & \textbf{Hindi} & \textbf{Sanskrit} & \textbf{Sanskrit Density} & \textbf{Sanskrit Compactness} \\ 
 & \textbf{Tokens} & \textbf{Tokens} & \textbf{Tokens} & \textbf{(CpT)} & \textbf{(vs. English)} \\ \midrule
SentencePiece (Baseline) & 41.96 & 43.09 & \textbf{21.67} & \textbf{5.07} & \textbf{0.52x (Most Compact)} \\
Gemini (Pro) & 44.44 & 52.10 & 48.33 & 2.65 & 1.09x \\
GPT o200k\_base & 45.62 & 62.28 & 53.70 & 2.38 & 1.18x \\
GPT cl100k\_base & 46.20 & 192.11 & 111.46 & 1.13 & 2.41x (Least Compact) \\ \bottomrule
\end{tabular}%
}
\end{table}

\begin{figure}[htbp]
    \centering
    \includegraphics[width=0.95\textwidth]{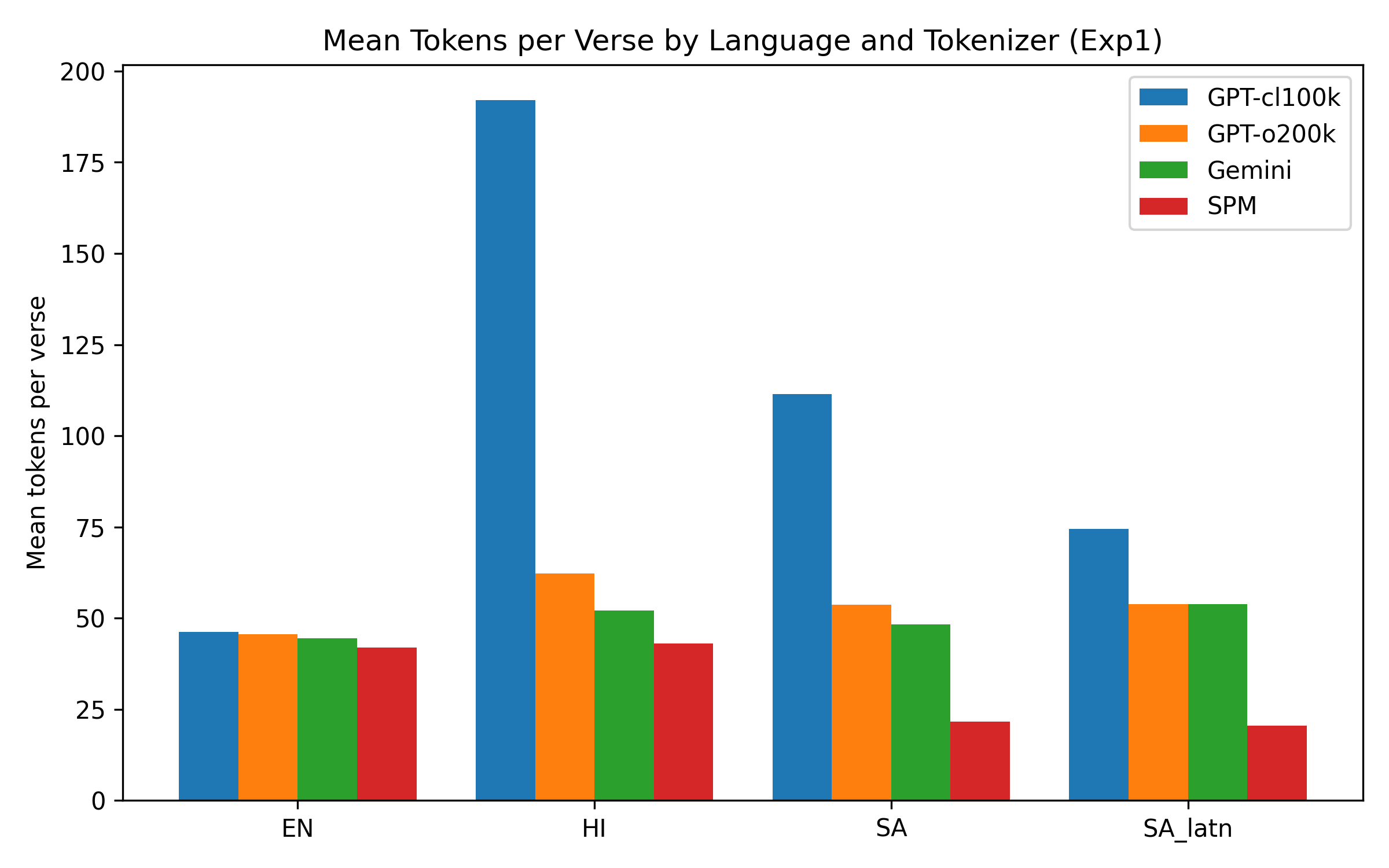}
    \vspace{0.3cm} 
    \includegraphics[width=0.95\textwidth]{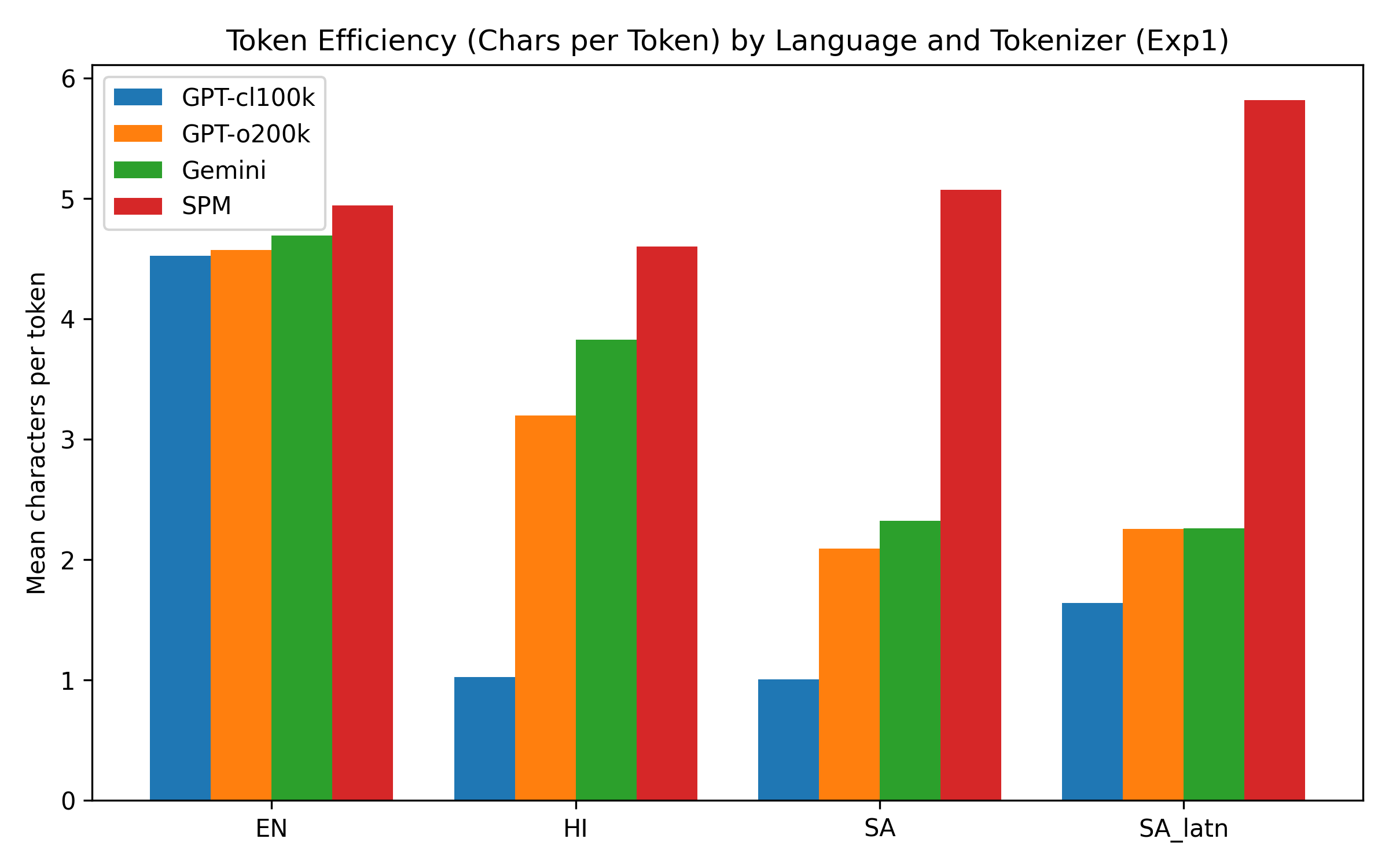}
    \caption{\textbf{Top:} Mean Tokens per Verse by Language and Tokenizer (Experiment 1). \textbf{Bottom:} Token Efficiency (Chars per Token) by Language and Tokenizer.}
    \label{fig:exp1}
\end{figure}

\subsection{Experiment 2: The Cost of Explanation}
Table \ref{tab:exp2} quantifies the semantic expansion—the additional token count required to translate a compact Sanskrit verse into explanatory English or Hindi commentary.

\begin{table}[ht]
\centering
\caption{Semantic Expansion Factor (Experiment 2)}
\label{tab:exp2}
\resizebox{\textwidth}{!}{%
\begin{tabular}{@{}lccccc@{}}
\toprule
\textbf{Tokenizer} & \textbf{Sanskrit Verse} & \textbf{English Commentary} & \textbf{Hindi Commentary} & \textbf{Expansion Factor} & \textbf{Expansion Factor} \\ 
 & \textbf{Tokens} & \textbf{Tokens} & \textbf{Tokens} & \textbf{(En Comm / Verse)} & \textbf{(Hi Comm / Verse)} \\ \midrule
SentencePiece & \textbf{21.67} & 428.34 & 416.75 & \textbf{19.8x} & \textbf{19.2x} \\
GPT o200k\_base & 53.70 & 374.27 & 506.50 & 7.0x & 9.4x \\
Gemini (Pro) & 48.33 & 368.59 & 421.64 & 7.6x & 8.7x \\
GPT cl100k\_base & 111.46 & 391.23 & 1589.30 & 3.5x* & 14.3x \\ \bottomrule
\end{tabular}%
}
\end{table}

\noindent \small{*Note: For \texttt{cl100k}, the English expansion factor (3.5x) is artificially low because the Sanskrit Verse token count is inflated by inefficient tokenization.}

\begin{figure}[htbp]
    \centering
    \includegraphics[width=0.95\textwidth]{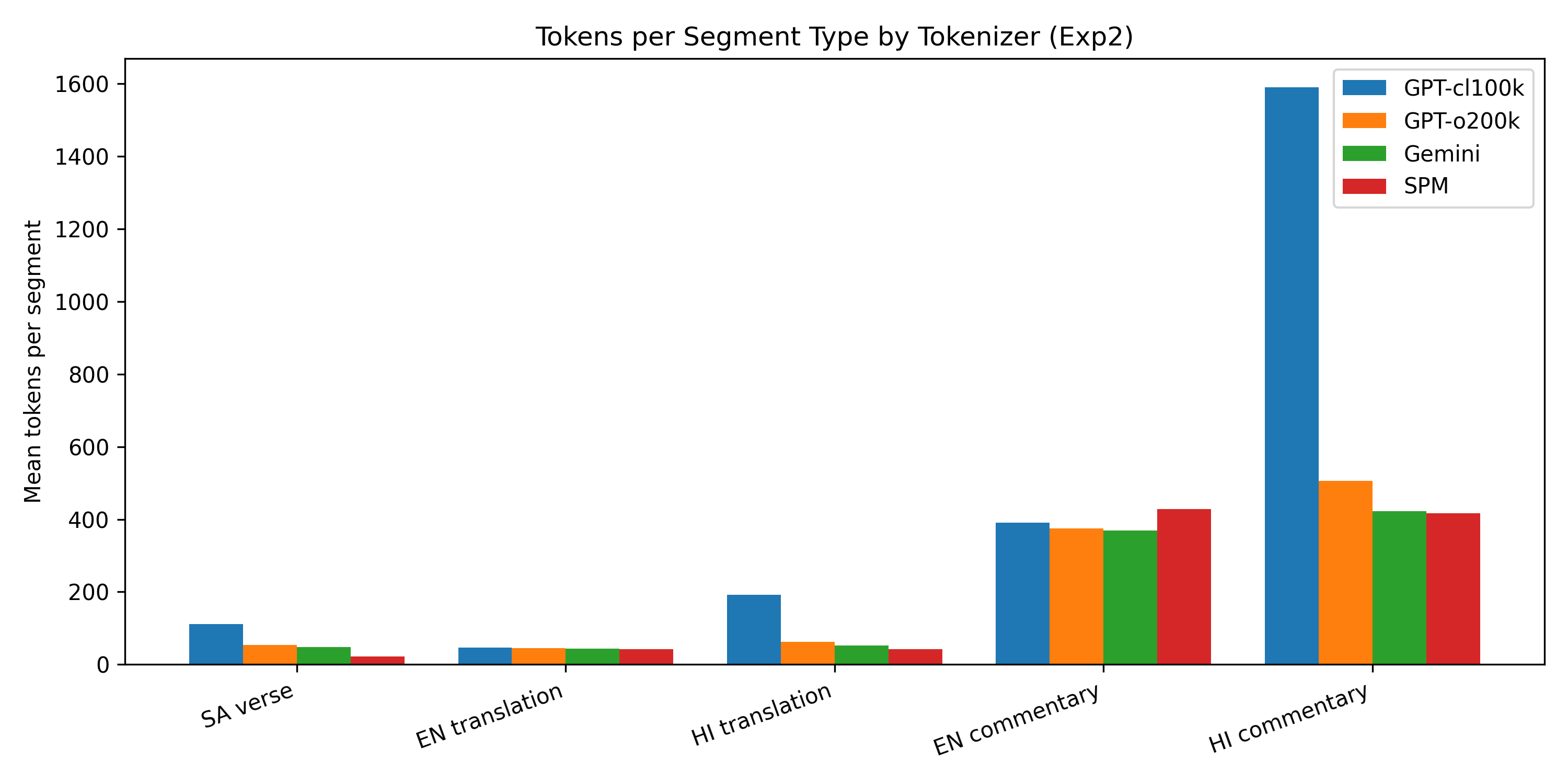}
    \vspace{0.3cm} 
    \includegraphics[width=0.95\textwidth]{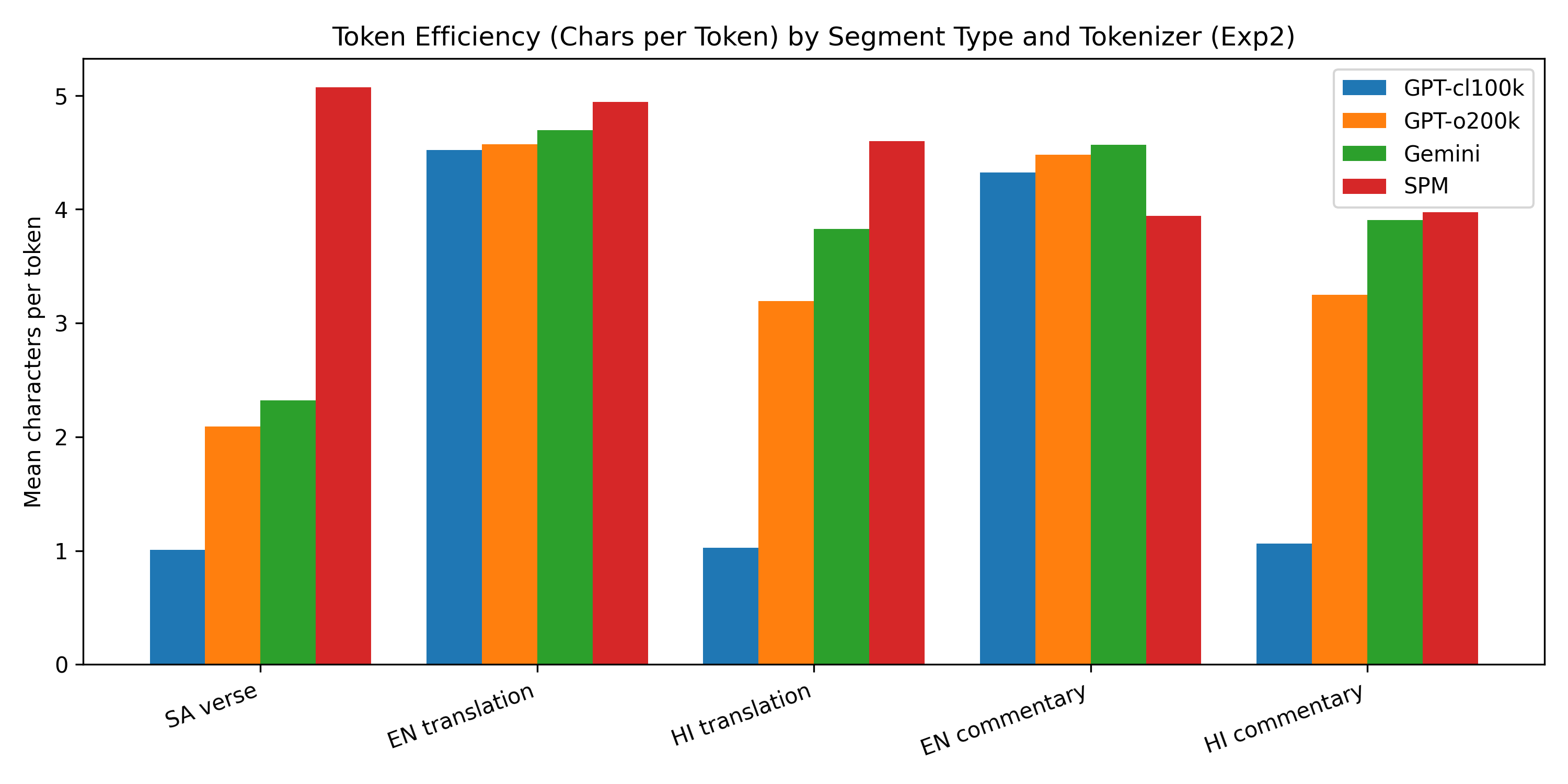}
    \caption{\textbf{Top:} Tokens per Segment Type by Tokenizer (Experiment 2). \textbf{Bottom:} Token Efficiency (Chars per Token) by Segment Type.}
    \label{fig:exp2}
\end{figure}

These results confirm that Sanskrit is among the most information-dense languages for digital encoding, capable of storing large semantic content with minimal token usage.

\FloatBarrier 

\section{Discussion}
The token efficiency observed in Sanskrit is due to the formalized grammar of 4,000 \textit{sutras} (generative rules) by Maharishi Panini, which encodes syntactic and semantic information through deterministic morphology rather than word ordering. This structure offers two distinct advantages for Large Language Models.

\paragraph{1. Algorithmic Compression via Sandhi}
Panini’s rule 6.1.77 states that if the vowels \textit{i, u, \d{r},} and \textit{\d{l}} in Sanskrit are followed by a dissimilar vowel, they are replaced by the semivowels \textit{y, v, r,} and \textit{l}.
\begin{itemize}
    \item \textbf{Input:} \textit{Iti} (End) + \textit{Adi} (Beginning)
    \item \textbf{Output:} \textit{Ityadi} (Deterministic and 100\% predictable).
\end{itemize}
Instead of memorizing patterns as in English, a Sanskrit-trained model can theoretically achieve higher accuracy with fewer parameters by learning these generative rules, effectively compressing the language model.

\paragraph{2. Information Density and the Vibhakti System}
Sanskrit is synthetic (agglutinative features), meaning a single word contains the Subject, Object, Mood, Verb, and Tense—information that typically requires a full sentence in English.
\begin{itemize}
    \item \textbf{English:} ``He wishes to go'' ($\sim$4 tokens).
    \item \textbf{Sanskrit:} ``\textit{Jigami\d{s}ati}'' ($\sim$1--2 tokens).
\end{itemize}
Furthermore, Sanskrit reduces dependency on positional embeddings due to its \textit{Vibhakti} system (independent word order). In English, ``Rama killed Ravana'' and ``Ravana killed Rama'' have vastly different meanings. In Sanskrit, the meaning is preserved regardless of order:
\begin{enumerate}
    \item \textit{Rāmaḥ Rāvaṇam hanti.}
    \item \textit{Rāvaṇam hanti Rāmaḥ.}
    \item \textit{Hanti Rāmaḥ Rāvaṇam.}
\end{enumerate}
In all cases, \textit{Rāmaḥ} (ending in -ḥ) marks the Subject and \textit{Rāvaṇam} (ending in -m) marks the Object. By explicitly encoding entity relationships within the word itself, Sanskrit reduces the computation required by the attention mechanism in Transformers, particularly for very long texts.

\section{Limitations}
\begin{itemize}
    \item In this study, we focus on the efficiency of tokenizers only and not the inference performance of the models. The findings reflect tokenizer behavior rather than the accuracy, fluency, or hallucination rates of the LLMs.
    \item The dataset is limited to the Bhagavad Gita, which may not represent the complete modern or classical Sanskrit language.
    \item English and Hindi translations vary slightly across sources.
    \item SentencePiece models were trained with an equal vocabulary size for fairness, but changes in parameters (e.g., vocabulary size) may yield different results.
\end{itemize}

\section{Future Work}
\paragraph{1. Linguistically Aware Tokenization}
The next step is the development of a grammar and morphology-aware tokenizer for Sanskrit. Unlike BPE, which relies on probabilistic frequency, a hybrid tokenizer could explicitly leverage Panini’s grammar rules—specifically \textit{Sandhi} (compound splitting) and \textit{Vibhakti} (inflectional parsing)—to achieve optimal compression. Such an architecture could theoretically surpass the efficiency of the SentencePiece baseline established in this study.

\paragraph{2. Native Sanskrit Foundation Models}
Our findings suggest that Sanskrit may be an ideal substrate for \textbf{efficient pre-training}. Future work should explore training a foundational Large Language Model (LLM) on a large Sanskrit corpus. If our hypothesis holds, such a model could encode vast amounts of knowledge with fewer parameters and lower inference costs than current English-centric models.

\section{Conclusion}
Our research confirms that Sanskrit is computationally more efficient than English and Hindi. However, this advantage is currently neglected by the design of commercial tokenizers. Under an unbiased baseline of SentencePiece, Sanskrit requires \textbf{$\sim$50\% fewer tokens} than its English counterpart to convey the same meaning. However, legacy systems like \texttt{cl100k\_base} invert this relationship, inflating Sanskrit’s token count by nearly \textbf{2.5x} compared to English. This discrepancy functions as a hidden ``token tax'' on Indic language users, artificially inflating cost.

Future research should not be limited to vocabulary adjustments but explore morphology-aware tokenization that integrates Panini’s \textit{sutras}. Furthermore, given the extreme storage density and rule-bound nature of Sanskrit, a pre-trained model trained on a large Sanskrit corpus could significantly reduce token usage, potentially lowering memory bandwidth and inference costs compared to today’s English-centric architectures.


\end{document}